\documentclass[letterpaper]{article} 

\usepackage{aaai25}  

\usepackage{times}  
\usepackage{helvet}  
\usepackage{courier}  
\usepackage[hyphens]{url}  
\usepackage{graphicx} 
\usepackage{natbib}  
\usepackage{caption} 
\usepackage{subcaption}
\usepackage[inkscapelatex=false]{svg}
\usepackage{algorithm}
\usepackage{algorithmic}

\usepackage{newfloat}
\usepackage{listings}
\DeclareCaptionStyle{ruled}{labelfont=normalfont,labelsep=colon,strut=off} 
\floatstyle{ruled}
\newfloat{listing}{tb}{lst}{}
\floatname{listing}{Listing}

\usepackage{amsmath}
\usepackage{amssymb}
\usepackage{subfig}
\usepackage{booktabs}

\nocopyright

\title{HMoE: Heterogeneous Mixture of Experts for Language Modeling}
\author{
    An Wang\equalcontrib\textsuperscript{,\rm 1},
    Xingwu Sun\equalcontrib\textsuperscript{,\rm 1},
    Ruobing Xie\textsuperscript{\rm 1,}\thanks{Corresponding Author}, Shuaipeng Li\textsuperscript{\rm 1} \\
    Jiaqi Zhu\textsuperscript{\rm 1}, Zhen Yang\textsuperscript{\rm 1}, Pinxue Zhao\textsuperscript{\rm 1}, J.N. Han\textsuperscript{\rm 1}, \\
    Zhanhui Kang\textsuperscript{\rm 1}, Di Wang\textsuperscript{\rm 1}, Naoaki Okazaki\textsuperscript{\rm 2}, Cheng-zhong Xu\textsuperscript{\rm 3}
}
\affiliations{
    \textsuperscript{\rm 1} Tencent Hunyuan\\
    \textsuperscript{\rm 2} Tokyo Institute of Technology \\
    \textsuperscript{\rm 3} University of Macau \\
}

\usepackage{bibentry}

\begin{document}

\maketitle

\begin{abstract}

Mixture of Experts (MoE) offers remarkable performance and computational efficiency by selectively activating subsets of model parameters. Traditionally, MoE models use homogeneous experts, each with identical capacity. However, varying complexity in input data necessitates experts with diverse capabilities, while homogeneous MoE hinders effective expert specialization and efficient parameter utilization. In this study, we propose a novel Heterogeneous Mixture of Experts (HMoE), where experts differ in size and thus possess diverse capacities. This heterogeneity allows for more specialized experts to handle varying token complexities more effectively. To address the imbalance in expert activation, we propose a novel training objective that encourages the frequent activation of smaller experts, enhancing computational efficiency and parameter utilization. Extensive experiments demonstrate that HMoE achieves lower loss with fewer activated parameters and outperforms conventional homogeneous MoE models on various pre-training evaluation benchmarks. Codes will be released upon acceptance.

\end{abstract}

\section{Introduction}

Mixture of Experts (MoE) ~\cite{adaptive_moe,smoe,gshard,switch_transformer,mixtral,deepseekmoe} is a cutting-edge technique in the field of large language models (LLMs) ~\cite{few_shot_learners,gpt4,rlhf,llama1,llama2,llama3} that excels in both performance and computational efficiency. At its core, MoE operates on the principle of dividing a model into multiple components, known as experts ~\cite{smoe}, each specializing in different tasks or aspects of the data. This specialization allows MoE to activate a subset of parameters, significantly enhancing the model's robustness and flexibility. The main advantage of MoE lies in that it can scale model parameters without the corresponding increase in computational cost. 

\begin{figure}[!htp]
\centering
\begin{subfigure}[b]{0.50\columnwidth}
    \centering
    \includegraphics[width=\linewidth]{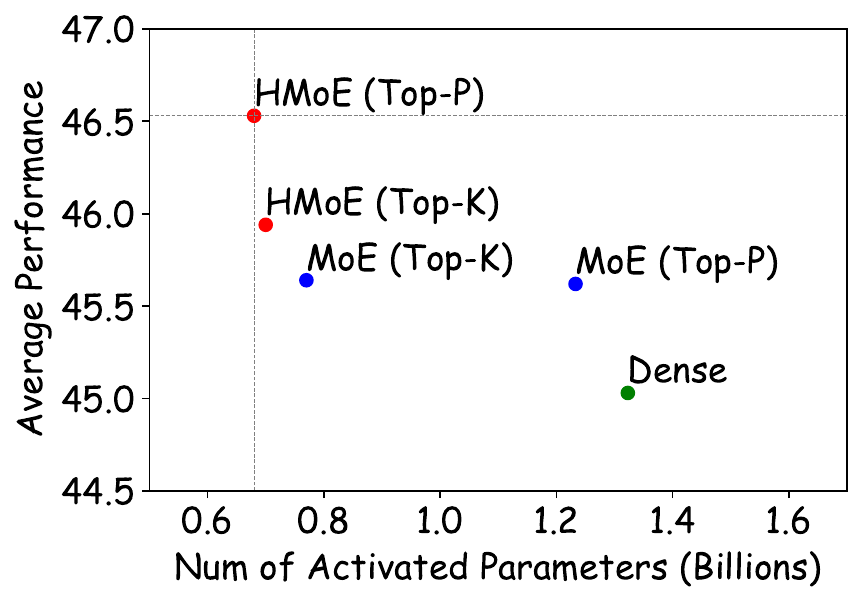}
    
\end{subfigure}
\hfill
\begin{subfigure}[b]{0.48\columnwidth}
    \centering
    \includegraphics[width=\linewidth]{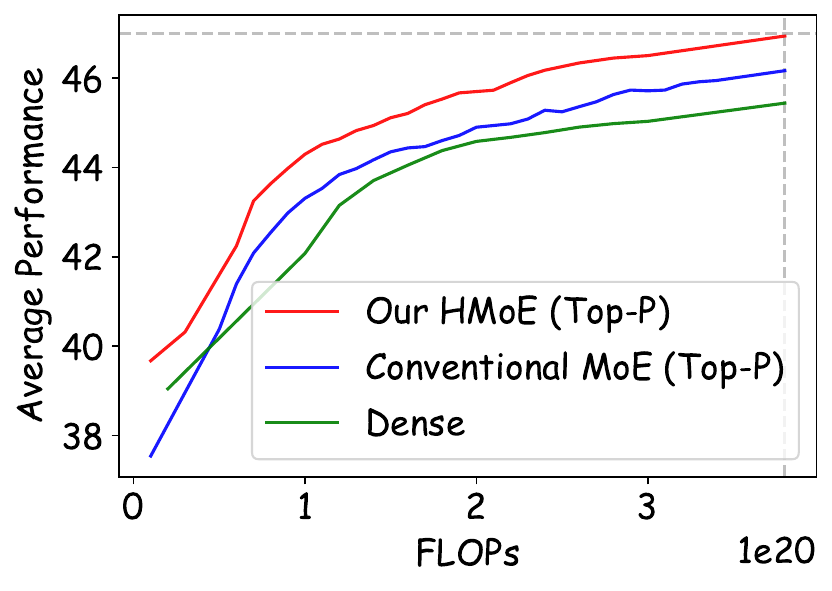}
    
\end{subfigure}
\caption{Comparisons of our heterogeneous MoE-3B with conventional homogeneous MoE-3B. Our proposed HMoE is superior on both performance and efficiency.}
\label{fig:intro_compare}
\end{figure}

Recently, almost all MoE models \cite{mixtral,deepseekmoe,multi_head_moe} predominantly adopt homogeneous experts for LLM, where all experts are structured identically with the same size. This uniformity inevitably leads to equivalent representational capacities among all experts. As a result, homogeneous experts often exhibit a convergence phenomenon \cite{expert_choice_routing}, where they learn similar representations over time, diminishing their uniqueness and specialization potential. The lack of diversity among experts becomes a significant bottleneck, particularly when handling inputs that require distinct representational capacities, ultimately hindering the model's overall performance and its ability to generalize across varied tasks.
Moreover, the equivalent representational capacity and professional ability of these homogeneous experts limit their functional differentiation, making it challenging to meet the varied complexity demands of different inputs or tokens in NLP tasks \cite{topp}. Consequently, MoE models struggle with suboptimal parameter utilization, as their identical experts may not provide the necessary depth or nuance for more complex inputs.

To address these challenges, a straightforward idea is to change the current homogeneous experts to heterogeneous ones. However, the challenges of heterogeneous MoE are mainly located in the following aspects: (a) \textbf{\emph{How to introduce appropriate heterogeneity to experts?}} This fundamental difference between homogeneous and heterogeneous MoE significantly impacts performance. (b) \textbf{\emph{How to design and guide the desired load distributions for heterogeneous experts?}} The optimal activation of heterogeneous experts is different from that in conventional MoE. We should first conclude what kind of expert activation distribution is optimal for heterogeneous MoE, and then provide effective guidance towards such activation, balancing both parameter efficiency and model effectiveness.

In this study, we propose a novel \textbf{Heterogeneous Mixture of Experts (HMoE)} structure as a pre-trained language model. Specifically, we empirically assign different sizes for experts to bring in heterogeneity. Our explorations reveal that such intuitive HMoE without any training guidance does not significantly surpass conventional MoE. During training, larger experts are overly activated, while smaller ones are underutilized. This imbalance activation results in a reduction in the model's representational capacity, which hinders the usage of heterogeneous experts.

Therefore, we propose a novel set of HMoE training objectives that \emph{encourages the activation of smaller experts}, leading to a more rational allocation of activated parameters and improved computational efficiency. 
Besides, we analyze three strategies of designing different heterogeneous expert size distributions, discovering the insights of \emph{optimal heterogeneity of experts in HMoE}.
Figure \ref{fig:intro_compare} demonstrates that our HMoE achieves better performance with fewer activated parameters, consistently outperforming traditional homogeneous MoE on pre-training evaluation benchmarks.
We conduct extensive experiments to verify the effectiveness and efficiency of our proposed HMoE, along with in-depth analyses. We contribute to the success of our enhanced HMoE for following reasons:
(a) Experts of varying sizes provide diverse capacities and promote higher specialization. (b) Expert heterogeneity ensures complex tokens get the necessary resources while simpler tokens are processed economically. (c) Leveraging MoE's inherent imbalance by activating more small experts to enhance their overall capability and further reduce computing costs.

We summarize the contributions of this work as follows:
\begin{itemize}
    \item 
    We introduce a novel HMoE model. It allows for enhanced specialization and a more granular response to diverse token complexities, improving both effectiveness and efficiency. To the best of our knowledge, this work is the first work exploring HMoE as a base language model.
    \item We propose a new set of training objectives that encourages the activation of smaller experts, leading to more efficient utilization of experts and preventing the disproportionate reliance on larger experts. We also explore different types of heterogeneity strategies for HMoE.
    \item Our experiments demonstrate that our HMoE achieves stronger performance with fewer activated parameters, thereby enhancing computational efficiency without sacrificing various downstream performances.
\end{itemize}

\section{Methodology}

\begin{figure*}[h]
\centering
\begin{subfigure}[b]{0.45\textwidth}
    \centering
    \includegraphics[width=0.99\linewidth]{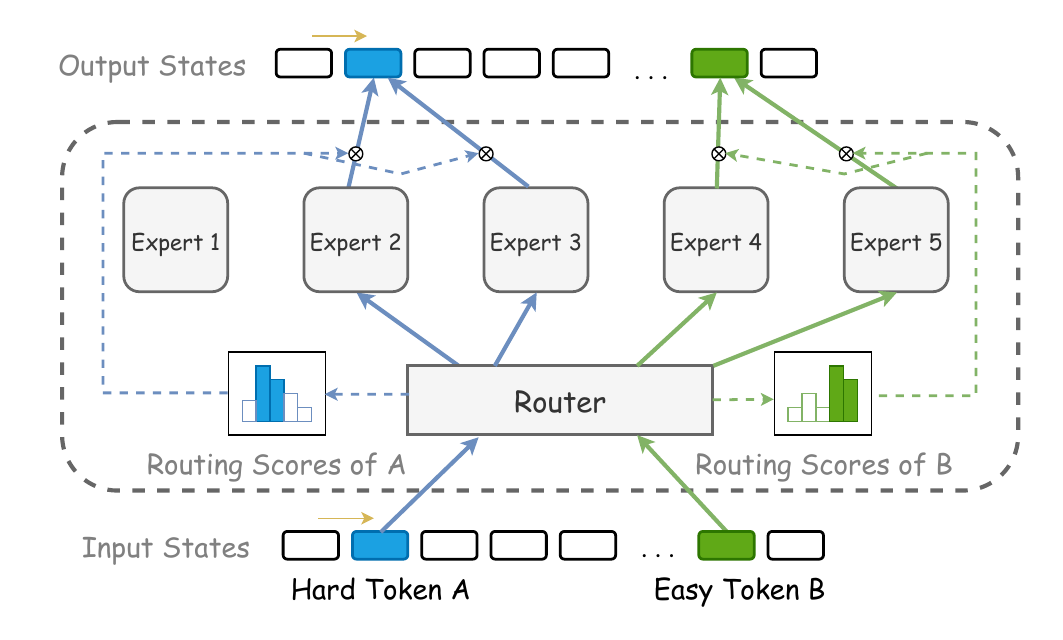}
    \caption{Conventional homogenerous MoE.}
\end{subfigure}
\begin{subfigure}[b]{0.53\textwidth}
    \centering
    \includegraphics[width=0.99\linewidth]{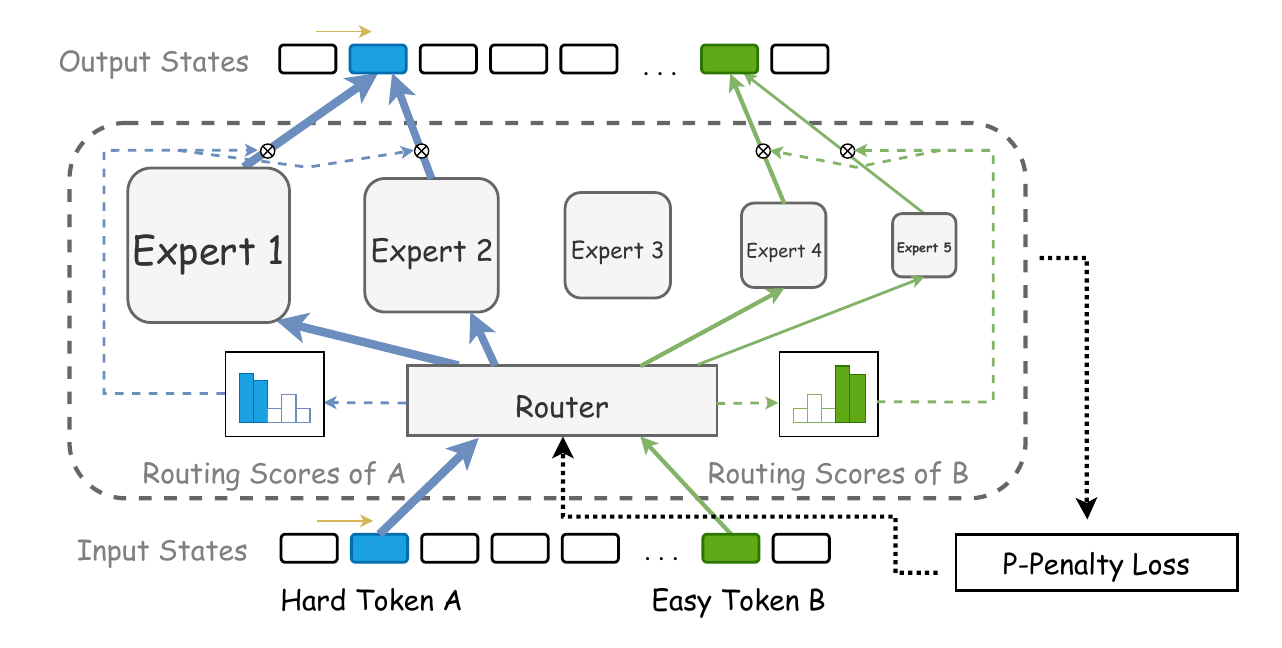}
    \caption{Our proposed heterogeneous MoE.}
\end{subfigure}
\caption{Two distinct model structures for Mixtures of Experts (MoE) are compared: (a) conventional homogeneous MoE model structure with all experts having identical parameter sizes; (b) our proposed heterogeneous MoE model structure characterized by substantial variations in parameter sizes of each expert, incorporating a parameter penalty loss during training to promote utilization of Experts with smaller parameter volumes. In our heterogeneous MoE, harder tokens are assigned to larger experts, while easier tokens are assigned to smaller experts. In conventional homogeneous MoE, all tokens are assigned to the same size experts regardless of their difficulty.}
\label{fig:model-architecture}
\end{figure*}

\subsection{Classical Mixture of Experts}

Different from dense models, most MoE models \cite{gshard,switch_transformer,topp,deepseekmoe,mixtral} replace the FFN layer of the transformer \cite{transformer} block with the MoE layer. The MoE layer consists of a router $g_i(\cdot)$ and multiple experts $\{e_1, e_2, ..., e_N\}$. The experts are composed of a set of independent Feed-Forward Network (FFN) layers. Experts are responsible for processing the input data according to their specialized knowledge. For each token, a subset of experts is activated to execute computations, and the router generates a probability distribution. The probability of this distribution indicates the likelihood of assigning the token to each expert.

\subsubsection{Routing Strategy}

The routing strategy is applied to select experts to be activated from $N$ experts. The \textbf{Top-K Routing}~\cite{smoe} strategy is the most widely-used strategy, which always activates a fixed number of experts for each token. It calculates the score which represents the probability of selecting each expert. We select the top $k$ experts with the highest scores to activate.

Recently, \textbf{Top-P Routing}~\cite{topp} is proposed to dynamically activate different numbers of experts for each token. Specifically, it first sorts scores from highest to lowest. Then given a fixed threshold $p$, if the highest probability is larger than the threshold, we only activate one expert. Otherwise, we progressively add additional experts until the cumulative probability exceeds the threshold $p$.

\subsubsection{Issues of Conventional Homogeneous MoE}

Currently, most work employs MoE layers in a homogeneous design. Each expert in the MoE layer usually has the same structure and size. Undoubtedly, this is a simple design that avoids introducing more hyperparameters. However, it also brings the following problems:

(1) \textbf{Lack of Expert Specialization}: Different experts within a homogeneous MoE show a tendency towards similarity \cite{expert_choice_routing}. Since homogeneous experts possess identical modeling capabilities, the router module randomly allocates tokens to these experts during pre-training. Consequently, without additional mechanisms to differentiate them, these experts might converge on similar features and patterns. As a result, the knowledge acquired by each expert lacks significant differentiation, leading to insufficient specialization among the experts.

(2) \textbf{Inefficient Parameter Allocation}: Most homogeneous MoE methods overlook the varying difficulties of tasks and the different complexities of tokens within the input. Smaller-sized experts can handle simpler tasks or easily understandable tokens effectively, while larger-sized experts are better suited for complex tasks and difficult tokens. However, homogeneous MoE models typically use experts of the same size for all inputs and tokens, leading to inefficient and suboptimal parameter allocation. The dynamic routing of Top-P Routing ~\cite{topp} attempts to address this issue by assigning different numbers of experts to different tokens. Nevertheless, it relies on fixed threshold settings and employs a rudimentary approach to difficulty modeling, making it challenging to adapt effectively to diverse inputs.

(3) \textbf{Representation Collapse and Load Imbalance}: Homogeneous MoE has a trend toward representation collapse \cite{representation_collaspe}. Representation collapse occurs when the majority of input tokens are assigned to only a few experts. This phenomenon also leads to load imbalance. The interconnected nature of representation collapse and load imbalance hampers the model's performance and efficiency.

\subsection{Exploration on Heterogeneous Mixture of Experts}

To alleviate the above issues in homogeneous MoE, we propose \textbf{Heterogeneous Mixture of Experts}. HMoE includes a router and expert network, with the key distinction that the models of experts within the same layer are different. To achieve an HMoE, we could design different structures and different sizes for experts. However, within the transformer model, experts with different structures make the training process extremely unstable. Therefore, in this work, we mainly explore HMoE with different expert sizes, as shown in Figure \ref{fig:model-architecture}.

\subsubsection{An Intuitive Exploration on HMoE}

For each expert $e_i$, we follow the FFN design in LLaMa ~\cite{llama1}. The detailed computation is as follows:
\begin{equation}
e_i(\mathbf{x}) = \mathbf{W}_{o,i} \cdot \left( \text{SiLU}(\mathbf{W}_{g,i} \cdot \mathbf{x} ) \odot (\mathbf{W}_{p,i} \cdot \mathbf{x} ) \right),
\end{equation}
\begin{equation}
\text{SiLU}(\mathbf{z}) = \mathbf{z} \cdot \sigma(\mathbf{z}), \quad \sigma(\mathbf{z}) = \frac{1}{1 + e^{-\mathbf{z}}},
\end{equation}
where $\mathbf{W}_{g,i} \in \mathbb{R}^{h_{\text{input}} \times h_{\text{ffn},i}}$, $\mathbf{W}_{p,i} \in \mathbb{R}^{h_{\text{input}} \times h_{\text{ffn},i}}$ and $\mathbf{W}_{o,i} \in \mathbb{R}^{h_{\text{ffn},i} \times h_{\text{input}}}$ are trainable parameters of expert $e_i$. $h_{\text{input}}$ and $h_{\text{ffn},i}$ are dim of input $x$ and hidden state in FFN.
To bring in heterogeneity for exploration, We intuitively change the hidden dim $h_{\text{ffn},i}$ to control the size of each expert $e_i$. 

\subsubsection{Results of Intuitive HMoE}

\begin{figure}[t]
\centering
\begin{subfigure}[b]{0.48\columnwidth} 
    \centering
    \includegraphics[width=\linewidth]{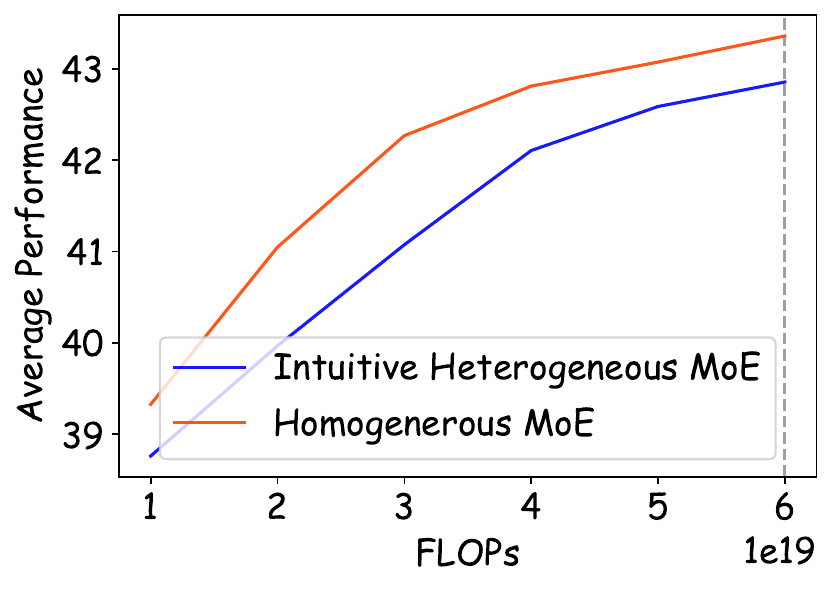} 
    
\end{subfigure}
\hfill 
\begin{subfigure}[b]{0.50\columnwidth} 
    \centering
    \includegraphics[width=\linewidth]{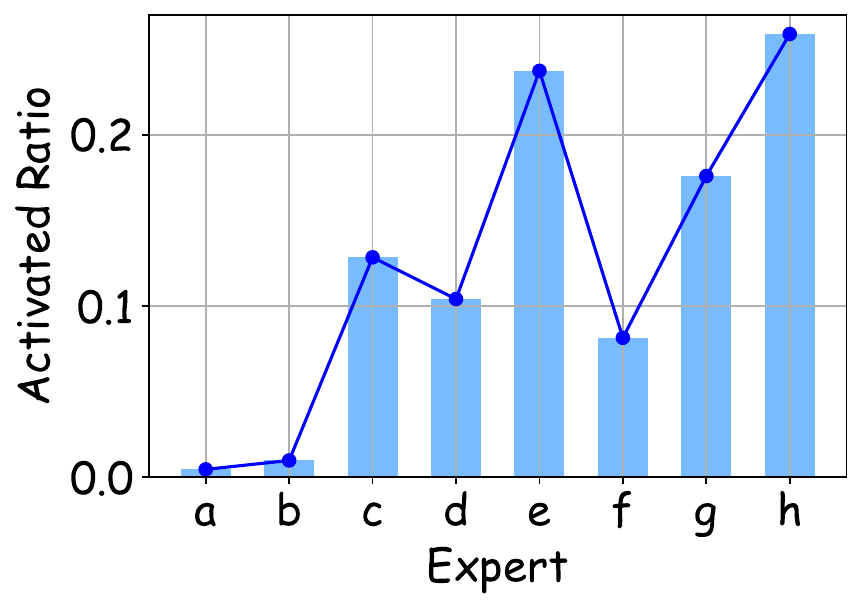} 
    
\end{subfigure}
\caption{Experimental results of intuitive exploration on HMoE. The left figure compares the performance of intuitive HMoE and conventional Homogeneous MoE. The Homogeneous MoE adapts load balancing loss while the intuitive Hetergeneous MoE does not utilize any auxiliary loss. The right figure shows the activated ratio of experts in the intuitive HMoE. The relative expert sizes in HMoE are $\{9,11,13,15,17,19,21,23\}$, matching experts \textit{a} to \textit{h}.}
\label{fig:intuitive_HMOE}
\end{figure}

We implement the aforementioned intuitive HMoE and conduct evaluation. Contrary to our expectations, the results do not demonstrate an improvement over the homogeneous MoE setup, as shown in Figure \ref{fig:intuitive_HMOE}.

Upon investigation, we discovered that the primary reason for this underperformance was the highly imbalanced load distribution among experts in the HMoE. Larger experts were activated more frequently, while smaller ones were rarely utilized. This imbalance led to a decline in the model's overall representational capacity. The root cause is that the larger experts possess stronger capabilities compared to the smaller ones, prompting the router to preferentially activate the larger experts more often. 

Nevertheless, we maintain that HMoE is still a very promising area of research because it has the potential to address the issue of \textbf{lack of expert specialization} by introducing diversity in the size and capacity of each expert. This inherent diversity allows the routing module to allocate tokens based on their complexity and characteristics, leading each expert to specialize in different aspects of the data. This mitigates the problem of experts converging towards similar representations and ensures that the model leverages a broader range of expertise.

\subsection{Enhanced Heterogeneous Mixture of Experts}

Considering the above-mentioned issues, we propose the following strategies to enhance HMoE.

\subsubsection{Activating More Small Experts}

In HMoE models, the presence of both large and small experts introduces a challenge where the optimization goal of the language model naturally favors the frequent activation of larger experts due to their superior performance. This tendency results in smaller experts being underutilized, while larger experts are activated more often, leading to a significant increase in activated parameters. This phenomenon diverges from the intended model objective, where we aim for larger experts to be primarily engaged in complex understanding and reasoning tasks, while smaller experts should be more universally applied to simpler tasks.

Previous research \cite{switch_transformer} adapts load balancing loss $ \mathcal{L}_{\text {lb}}$ to eliminate load unbalancing among different experts in Homogeneous MoE:
\begin{equation}
\begin{aligned}
& \mathcal{L}_{\text{lb}}=N \sum_{i=1}^N \mathcal{T}_i * \hat{\mathcal{P}}_i, \\
& \mathcal{T}_i=\frac{1}{T} \sum_{t=1}^{T} \text{1}\{ e_i \in E^t \}, \quad \hat{\mathcal{P}}_i=\frac{1}{T} \sum_{t=1}^{T} P_{i,t},
\end{aligned}
\end{equation}
where $\mathcal{T}_i$ represents the partation of tokens assigned to expert $e_i$. 
$\hat{\mathcal{P}}_i$ represents the gating probability assigned to $e_i$. 
$P_{i,t}$ represents the gating probability assigned to $e_i$ for token $x_t$. 
$E^t$ represents the set of activated experts for the token $x_t$.

The objective of the load balancing loss is to achieve experts evenly activated. Nevertheless, it does not satisfy our motivation for designing HMoE. Because of the disparity in expert sizes, the load-balancing loss fails to stop the model from preferring to activate larger experts. To address the issue where larger experts are predominantly utilized, leading to the underutilization of smaller experts and a considerable rise in activated parameters, we introduce a novel training objective \textbf{parameter penalty (P-Penalty) loss} $\mathcal{L}_{\text{P-Penalty}}$ as:
\begin{equation}
\begin{aligned}
& \mathcal{L}_{\text{P-Penalty}}=N \sum_{i=1}^N \mathcal{M}_i * \hat{\mathcal{P}}_i, \\
& \mathcal{M}_i=\frac{1}{T} \sum_{t=1}^{T} \text{1}\{ e_i \in E^t \} \times h_{\text{ffn},i}. \\
\end{aligned}
\end{equation}
$\mathcal{M}_i$ represents the average dimension of the hidden state of the expert $e_i$ on the entire input $x$. It imports the influence of expert size into the loss.
When the model employs more large experts, the loss rises. Hence, it will direct the model to more economically utilize smaller experts. In contrast, for harder tasks, using larger experts can yield greater benefits than parameter penalties. At this point, larger experts will also be activated to take part in the calculation. To be noted, if all expert has the same size, our parameter penalty loss is equal to the classical load balancing loss.

Besides, with the Top-P routing strategy, we find that MoE tends to activate an increasing number of experts during training, which reduces the efficiency of MoE. Therefore, we implement the router entropy loss \cite{topp} to prevent the model from using too many parameters, maintaining its ability to selectively activate experts as follows:
\begin{equation}
\begin{aligned}
& \mathcal{L}_{\text{entropy}}=N \sum_{i=1}^N \text{P}_i \times \text{log}(\text{P}_i).
\end{aligned}
\end{equation}
In our HMoE, besides the original \emph{language modeling loss}, the final loss for both Top-K and Top-P routing strategies further includes the \emph{parameter penalty loss} $\mathcal{L}_{\text{P-Penalty}}$, with Top-P additionally incorporating the \emph{router entropy loss} $\mathcal{L}_{\text{entropy}}$.

\begin{table*}[t]
\centering
\small
\begin{tabular}{lcccccccc}
\toprule
Method & Activated Parameters & PIQA & hellaswag & BoolQ & ARC-Easy & winogrande & SIQA & AVG \\
\midrule
\multicolumn{8}{c}{ $7 \times 10^{19}$ FLOPs Training } \\
Dense-0.4B & 417M & 55.55 & 26.33 & 57.90 & 30.88 & 51.38 & 32.80 & 42.47 \\
MoE-0.4B (Top-K)& 163M & 57.67 & 27.81 & 62.13 & 29.70 & 50.59 & 32.82 & 43.45 \\
MoE-0.4B (Top-P) & 173M & 56.92 & 27.73 & 56.54 & 30.18 & 51.67 & 32.89 & 42.66  \\
HMoE-0.4B (Top-K) & 153M & 56.67 & \textbf{28.26} & 59.80 & 31.93 & \textbf{52.49} & \textbf{32.91} & 43.68 \\
HMoE-0.4B (Top-P) & 173M & \textbf{58.98} & 28.10 & \textbf{60.78} & \textbf{34.14} & 52.21 & 32.83 & \textbf{44.51} \\
\midrule
\multicolumn{8}{c}{ $2.6 \times 10^{20}$ FLOPs Training} \\
Dense-1B & 1.32B & 58.92 & 29.57 & \textbf{61.70} & 35.26 & 51.85 & 32.86 & 45.03\\
MoE-3B (Top-K) & 0.77B & \textbf{61.92} & 32.80 & 60.06 &  33.96 & 52.51 & 32.58 & 45.64\\
MoE-3B (Top-P) & 1.23B & 61.42 & 32.16 & 61.47 & 33.51 & 52.27 &  32.91 & 45.62\\
HMoE-3B (Top-K) & 0.70B & 61.04 & 32.89 & 60.26 & 36.14 & 52.49 &  32.82 & 45.94\\
HMoE-3B (Top-P) & 0.68B & 61.79 & \textbf{33.22} &  61.69 & \textbf{36.49} & \textbf{52.96} &  \textbf{33.00} & \textbf{46.53}\\
\bottomrule
\end{tabular}
\caption{Results on pre-training model evaluation benchmarks. Our HMoE consistently outperforms Homogenerous MoE.}
\label{table:main_result}
\end{table*}

\subsubsection{Designing More Optimal Heterogeneity for Experts}

Intuitively, the specific sizes of each heterogeneous expert have a large impact on the final results. In this work, we mainly explore three types of heterogeneity structures:

(1) \textbf{\emph{Geometric strategy}}. This strategy assigns the distribution of expert sizes following a geometric sequence. For example, we configure the relative size proportions of the experts to be $\{1, 2, 4, 8, 16, 32, 64, 128\}$ as in the intuitive exploration. It has a relatively high level of heterogeneity of experts. As a result, it highlights key experts, allowing them to play a more significant role in computation. More computing resources are allocated to larger-scale experts when dealing with complex and important tasks. However, it inevitably leads to an unbalanced resource allocation, where smaller-scale experts might be overly neglected in most cases. Therefore, this design may lead to severe load unbalancing. It might also be less applicable to tasks that require balanced handling of various possibilities, as it may overly emphasize certain situations.

(2) \textbf{\emph{Arithmetic strategy}}.
The distribution can also follow an arithmetic sequence (i.e., the size gap between adjacent experts is constant). For example, we set the relative expert size as $\{9,11,13,15,17,19,21,23\}$. The benefits of this strategy include a relatively balanced resource allocation and consistent variation in differences between experts. Compared with geometric progression, the difference between the largest and smallest experts in arithmetic progression is smaller, which makes even small experts have certain expressive abilities. Thus the strategy makes model training more stable. In this study, we mainly adapt this strategy for research HMoE.

(3) \textbf{\emph{Hybrid strategy}}.
The hybrid strategy that jointly combines both homogeneous and heterogeneous such as $\{1, 1, 1, 1, 2, 2, 4, 4\}$ is also a good competitor. We designed this setup based on the assumption that the MoE model requires multiple experts with similar capabilities or functionalities. Especially in scenarios involving expert combinations, completely differentiated experts might have drawbacks.
It has the flexibility to adjust the proportion of homogeneous and heterogeneous parts based on different task requirements.

As a pioneer of the exploration of HMoE, we propose three strategies of different heterogeneity levels and conduct extensive evaluations on different settings for more insights. More optimal HMoE distributions and structures will be explored in the future.

\section{Experiments}

\begin{figure}[t]
    \centering
\begin{subfigure}[b]{0.50\columnwidth} 
    \centering
    \includegraphics[width=\linewidth]{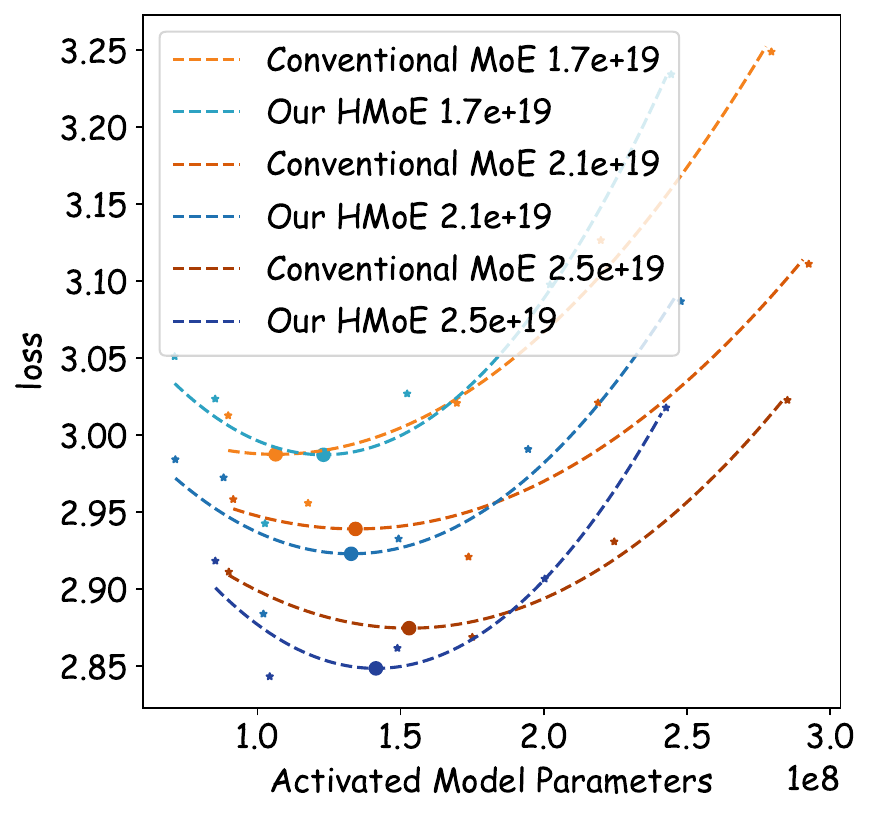} 
\end{subfigure}
\hfill 
\begin{subfigure}[b]{0.49\columnwidth} 
    \centering
    \includegraphics[width=\linewidth]{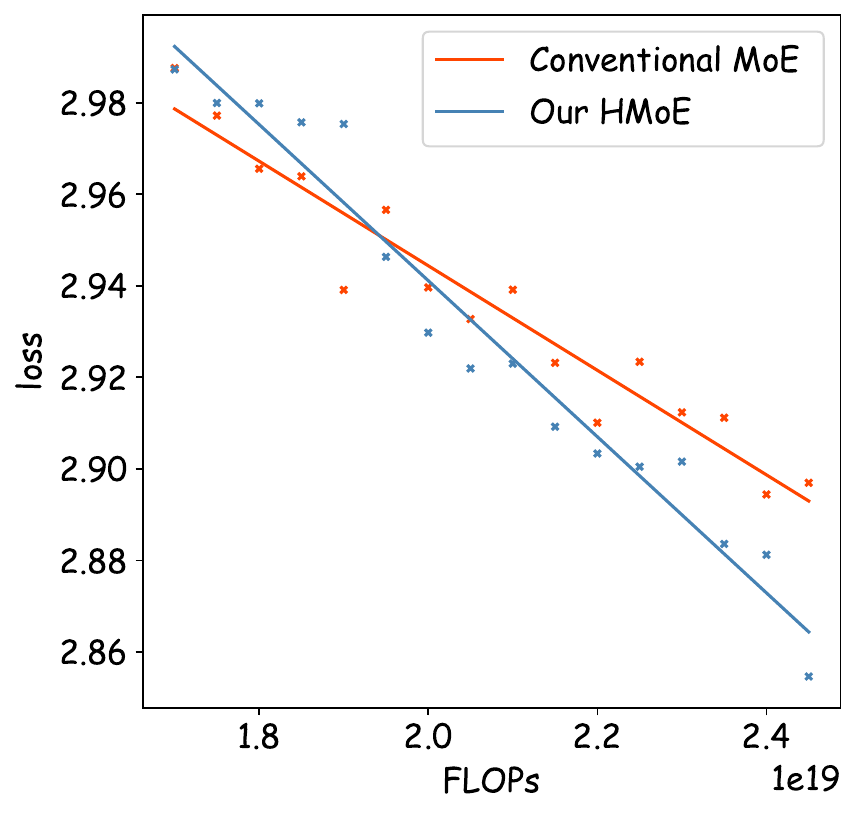}
\end{subfigure}    
    \caption{Analysis of isoFLOP for conventional MoE (Top-P) and our poposed HMoE (Top-P). The left figure depicts the optimal activated model parameters for various FLOPs. The right figure illustrates the variations in loss as FLOPs increase, given the optimal settings.}
    \label{fig:isoflop}
\end{figure}

\subsection{Experimental Settings}

\subsubsection{Pre-training Datasets}
For our pre-training data, we utilize the RedPajama ~\cite{redpajama} dataset. It is an open-source dataset consisting of various sources like the common crawl, C4~\cite{c4}, GitHub, Wikipedia, books~\cite{pile}, arXiv, and StackExchange.

\subsubsection{Competitors}

In our main experiment, we evaluate two types of baseline methods and our HMoE model: (1) \textbf{Dense}, which are standard Transformer decoder-only models without MoE layers, implemented with 0.4B and 1B parameters. (2) Homogeneous \textbf{MoE}, where FFN layers are replaced with MoE Layers including eight homogeneous experts, implemented with 0.4B and 3B total parameters, using both Top-K (k=2) and Top-P (p=0.6) routing strategies. (3) \textbf{HMoE}, our proposed method with Heterogeneous MoE Layers replacing FFN layers, also implemented with 0.4B and 3B parameters with both Top-K (k=2) and Top-P (p=0.6) strategies. To reflect the difference in performance between pure heterogeneous models and conventional homogeneous models, the expert size distribution employs an arithmetic strategy (The relative expert sizes are $\{9,11,13,15,17,19,21,23\}$).
The detailed setting is introduced in the Appendix.

\subsubsection{Evaluation}

We evaluate these models on six different benchmarks ~\cite{llm_eval} including PIQA ~\cite{piqa}, hellaswag~\cite{hellaswag}, BoolQ~\cite{boolq}, ARC~\cite{arc}, winogrande~\cite{winogrande} and SIQA~\cite{siqa}. These tasks examine models' language understanding, logical reasoning, knowledge utilization, and social awareness capabilities.
Since the activated parameters of different methods are varied, we ensure a fair comparison by basing our model evaluations on identical computational training costs (FLOPs) instead of the number of training tokens.

\subsection{Main Results}

Table \ref{table:main_result} presents a comparative analysis of the performance of various models on pre-training evaluation benchmarks.

(1) The results demonstrate the superiority of the MoE models over the Dense models across the board. Notably, our proposed HMoE models, utilizing both Top-K and Top-P routing strategies, have outperformed their traditional MoE and Dense counterparts in almost all evaluated metrics.

(2) Specifically, within the category of models utilizing $7 \times 10^{19}$ FLOPs, HMoE-0.4B model demonstrates a significant advantage, particularly with the Top-P routing strategy, surpassing Dense-0.4B model by an average of 2.04\

(3) When we shift our focus to models trained with a higher budget of $2.6 \times 10^{20}$ FLOPs, the HMoE-3B model with Top-P routing once again emerges as the top performer, outperforming the Dense-1B model by an average of 1.50\

(4) Furthermore, the comparison between Top-K and Top-P routing within the HMoE model is also insightful. The Top-P routing strategy generally yields better results, implying that the dynamic routing strategy cooperates well with heterogeneous experts. We attribute this to the fact that both Top-P routing and heterogeneous experts are designed to adapt to the complexity of the input.

We additionally conduct isoFLOP comparisons as shown in Figure \ref{fig:isoflop}. We found that due to expert heterogeneity, if the training FLOPs are too few, the performance of HMoE is not significantly superior to traditional MoE. However, at relatively early stages of training (around $2\times10^{19}$ FLOPs), HMoE already shows a stable trend of outperforming its homogeneous counterpart. It can be expected that with larger models and more data, the advantages of heterogeneity will become even more pronounced.

\subsection{Efficiency Analyses on HMoE}

\begin{figure}[t]
    \centering
\begin{subfigure}[b]{0.51\columnwidth} 
    \centering
    \includegraphics[width=\linewidth]
    {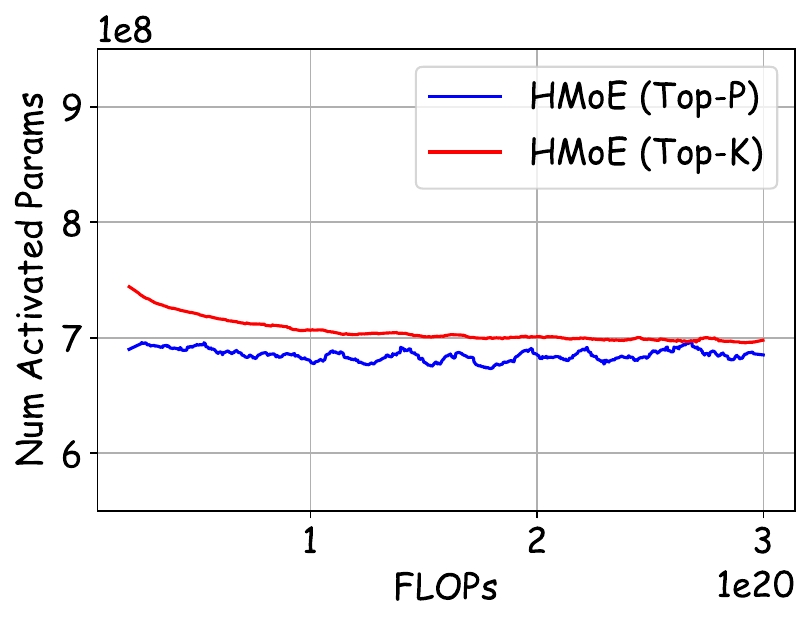} 
    
\end{subfigure}
\hfill 
\begin{subfigure}[b]{0.48\columnwidth} 
    \centering
    \includegraphics[width=\linewidth]{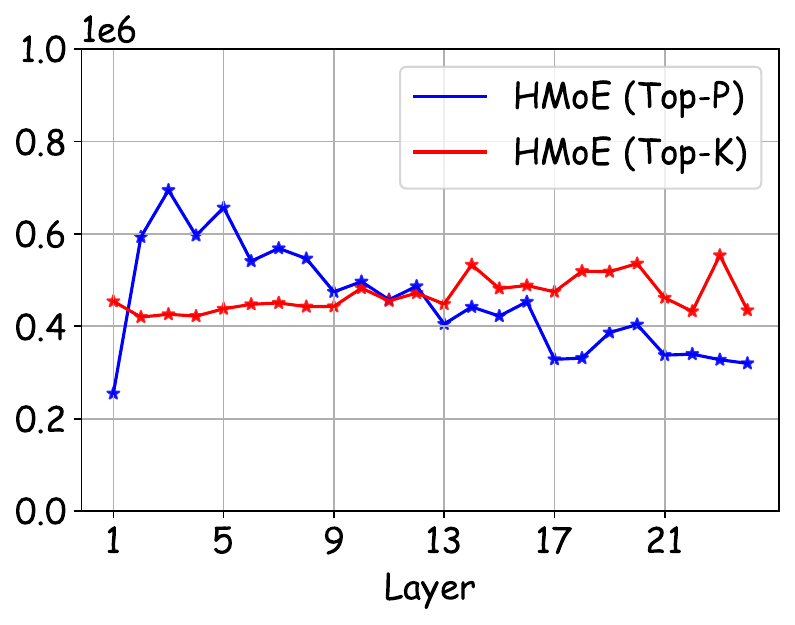}
    
\end{subfigure}    
    
    \caption{Average activated parameters across training FLOPs (left) or different layers (right).}
    \label{fig:avg-activated-parameters-change}
\end{figure}

\noindent
\textbf{Activated parameters of different MoE models}
The left side of figure \ref{fig:avg-activated-parameters-change} shows the average activated parameters during training. 
For HMoE models using Top-P and Top-K routing, the number of activated parameters stays stable and shows a downward trend over time. This is beneficial for large model training, keeping the HMoE's expected sparse activation property, even with more tokens. It is to be noted, that activation parameters for HMoE models are more stable with Top-K routing than with Top-P routing.

\begin{figure}[t]
    \centering
    \includegraphics[width=0.9\columnwidth]{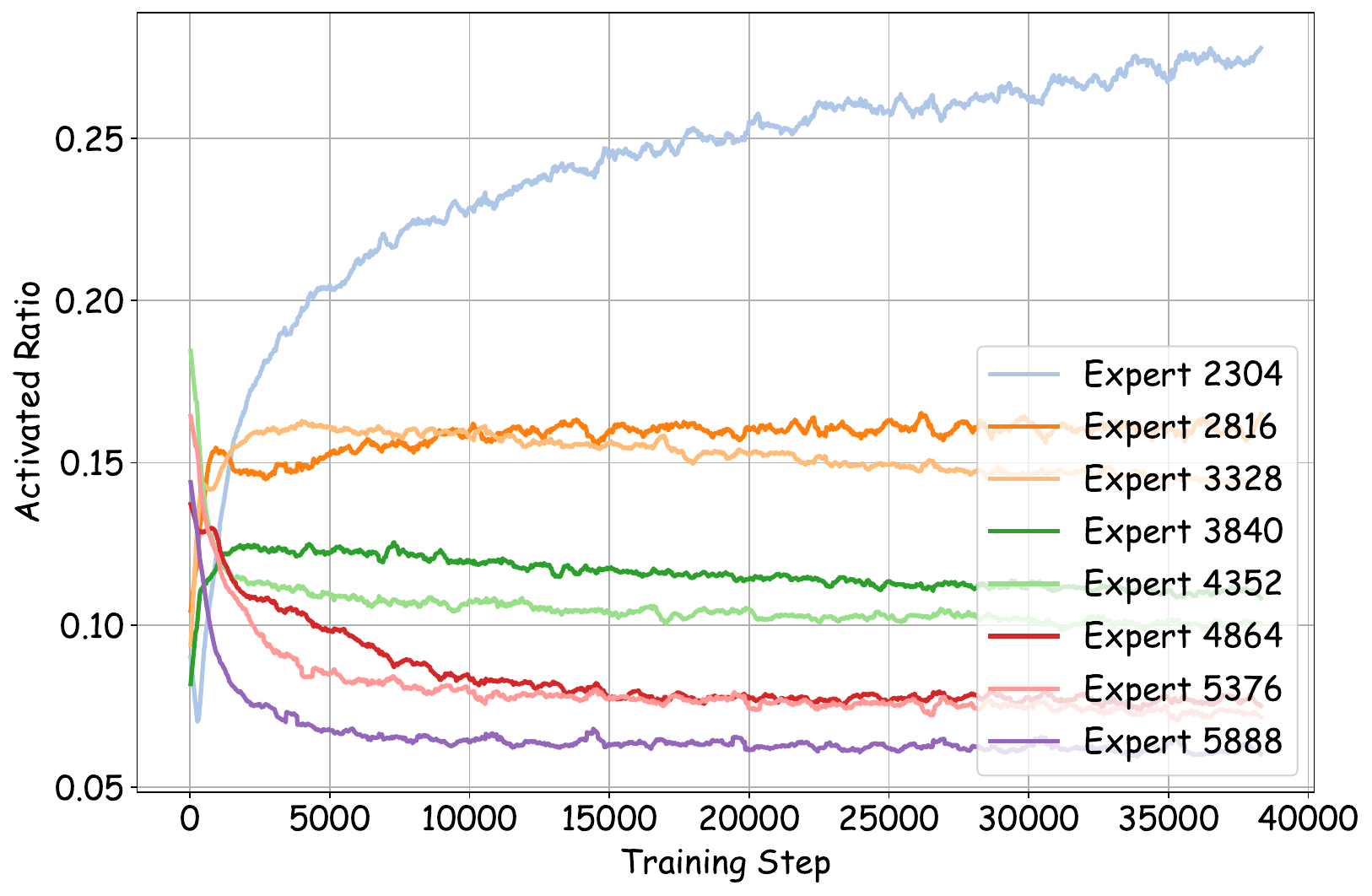}
    \caption{Activated parameters of experts in HMoE (Top-P). The values in the legend indicate the hidden dimensions of the experts, which represent their sizes.}
    \label{fig:expert-activated-parameters-change}
\end{figure}

\noindent
\textbf{Activated parameters of different experts in HMoE}
We explore the underlying causes of the stable or declining trend in activated parameters within HMoE with Top-P routing. As depicted in Figure \ref{fig:expert-activated-parameters-change}, the activation of smaller experts increases over the course of training, while larger experts experience a decline in their activation rates. This highlights the effectiveness of our proposed P-Penalty loss. The increased activation rates of smaller experts enhance their capacity to comprehend general knowledge, as further evidenced in Section \ref{expert_analysis}. This shift causes the role of smaller experts to increasingly resemble that of shared experts~\cite{deepseekmoe}. Additionally, the activation frequency of different experts remains constant throughout the training process, indicating the router's consistent token allocation.

\noindent
\textbf{Activated parameters of different layers in HMoE}
The right side of Figure \ref{fig:avg-activated-parameters-change} shows the layer-wise distribution of activated parameters. With Top-P routing, activated parameters decrease with layer depth. The first layer of HMoE with Top-P has a very low activation rate because nearly all tokens are routed to one expert, unlike other layers where activation is more balanced.

\subsection{Ablation Study}

\subsubsection{Effectiveness of Auxiliary Losses}

\begin{figure}[t]
    \centering
    \begin{subfigure}[b]{0.57\columnwidth}
        \centering
        \raisebox{-1cm}{ 
        \includegraphics[width=\linewidth]{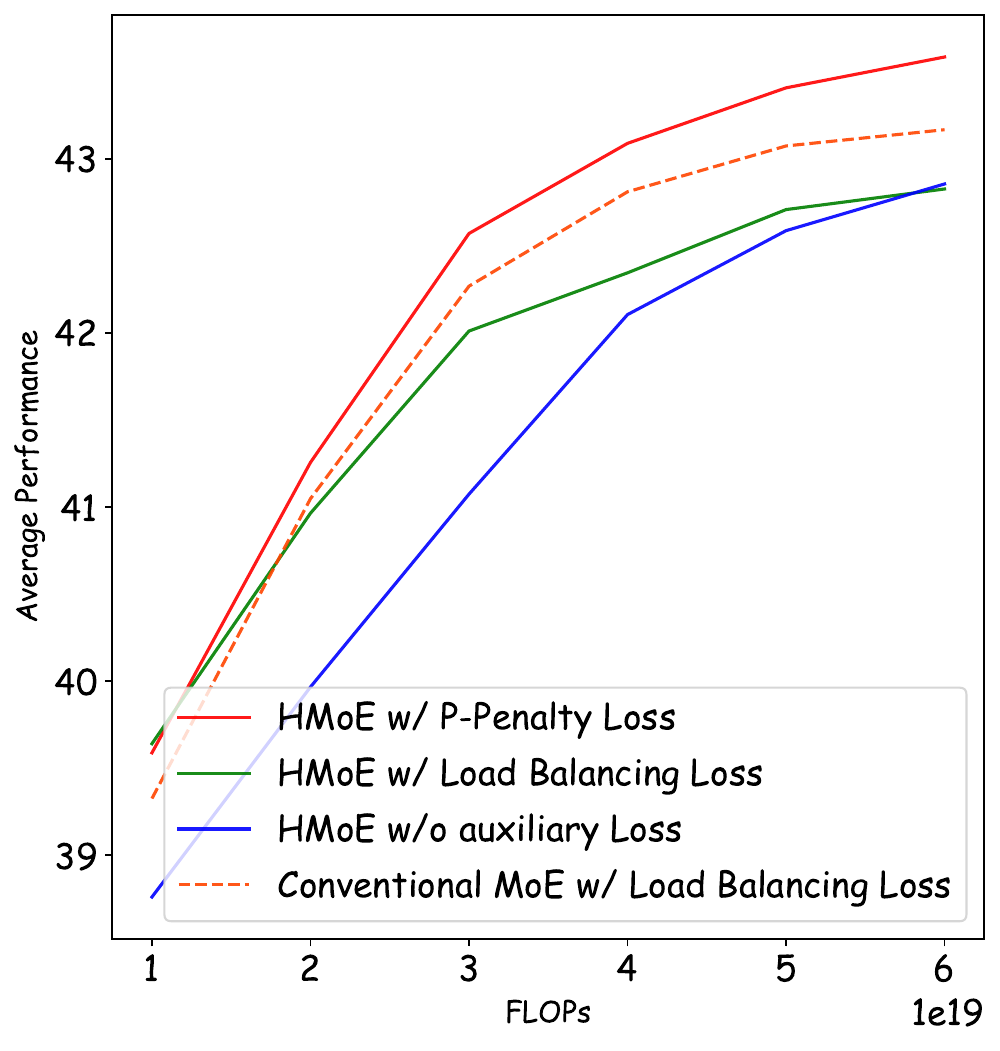}
        }
        \label{fig:loss_ablation}
        
    \end{subfigure}
    \hfill
    \begin{subfigure}[b]{0.42\columnwidth}
        \centering
        \begin{subfigure}[b]{\linewidth}
            \centering
            \includegraphics[width=\linewidth]{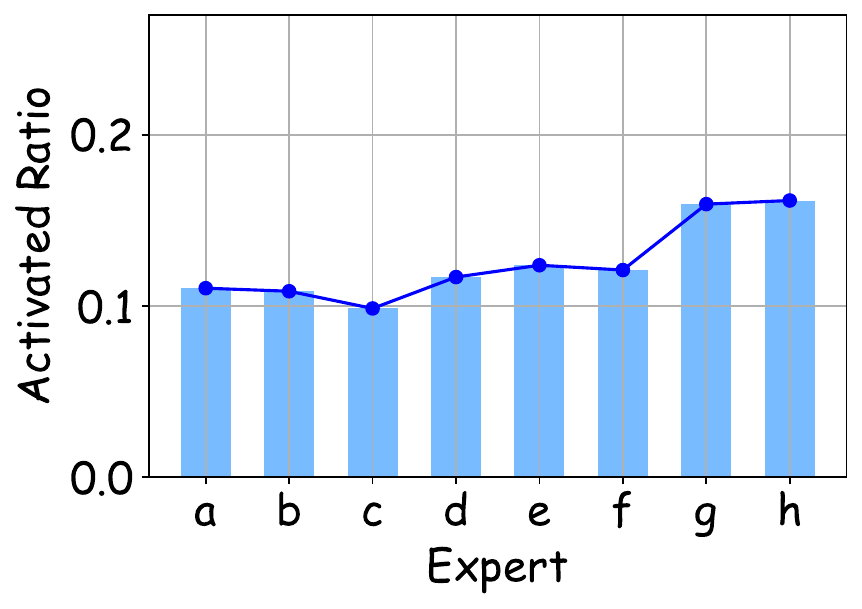}
            
        \end{subfigure}
        \vfill
        \begin{subfigure}[b]{\linewidth}
            \centering
            \includegraphics[width=\linewidth]{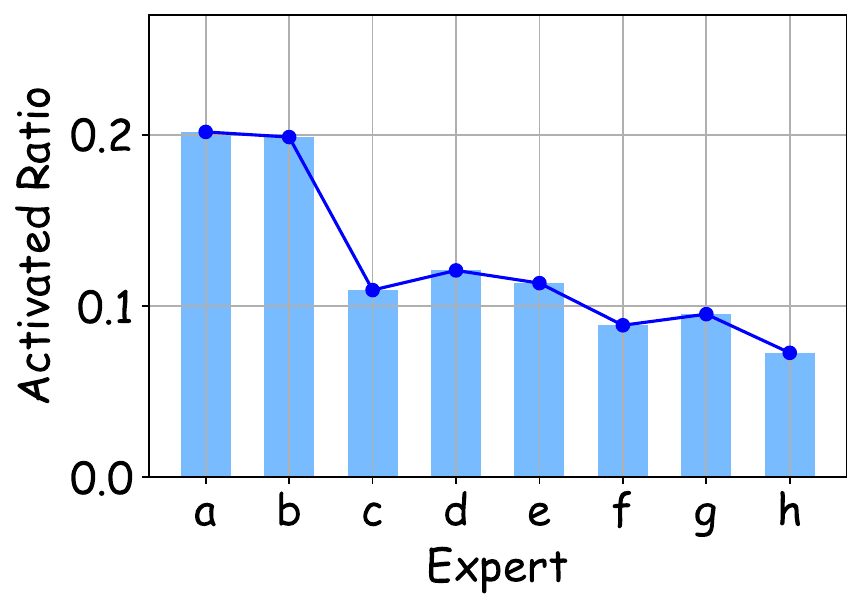}
            
        \end{subfigure}
    \end{subfigure}
    \caption{The left figure shows the effectiveness of auxiliary losses. The right figure shows the activated parameter ratio varying by model size across load balancing loss (above subfigure) and P-Penalty loss (below subfigure).}
    \label{fig:combined_loss_ablation}
\end{figure}

Our proposed P-Penalty loss is crucial for HMoE's performance. We conduct an ablation study to evaluate auxiliary losses.
As shown in Figure \ref{fig:combined_loss_ablation} (left), the P-Penalty loss yields the best results. Figures \ref{fig:intuitive_HMOE} (right) and \ref{fig:combined_loss_ablation} (right) illustrate the impact of auxiliary losses on expert activation. Although the load balancing loss fails to reduce the frequent activation of large experts, the P-Penalty loss successfully adjusts the model's goals to favor the activation of smaller experts more often, thereby greatly improving model performance.

\subsubsection{Analyses on Expert Heterogeneity}

\begin{figure}[t]
    \centering
    \begin{subfigure}[b]{0.49\columnwidth} 
    \centering
    \includegraphics[width=\linewidth]{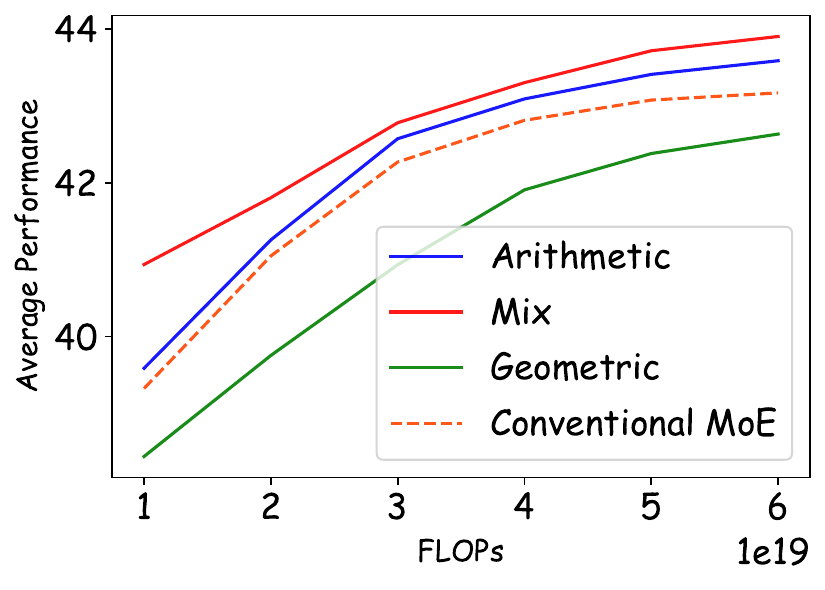} 
    
\end{subfigure}
\hfill 
\begin{subfigure}[b]{0.50\columnwidth} 
    \centering
    \includegraphics[width=\linewidth]{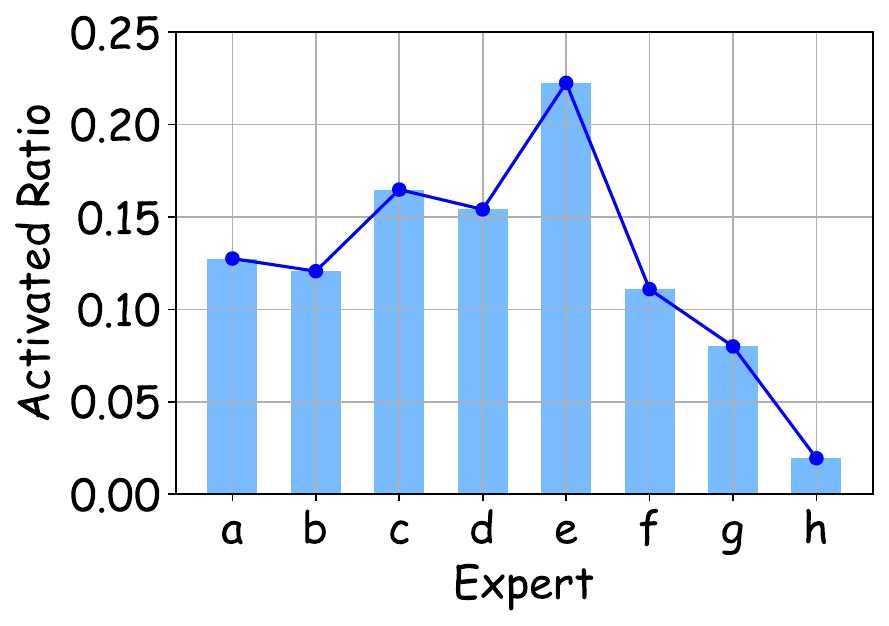}
    
\end{subfigure} 
    \caption{Analysis of expert heterogeneity through ablation. The figure on the left illustrates a performance comparison across various expert-size design strategies. The right figure displays the activation ratios of experts in HMoE using a \textbf{geometric} strategy.}
    \label{fig:expert_dist_ablation}
\end{figure}

The expert size distribution in HMoE significantly influences model performance. Figure \ref{fig:expert_dist_ablation} (left) compares HMoE across various distributions: geometric, arithmetic, and hybrid. Our results show that the geometric distribution performs the worst. Figure \ref{fig:expert_dist_ablation} (right) illustrates that smaller experts in the geometric progression are less frequently activated, even with P-Penalty loss, suggesting their capacity is insufficient because of their too-small size. Conversely, the hybrid model outperforms the arithmetic one, indicating that a mix of similar and varied expert sizes optimizes the HMoE model. This indicates that a mix of experts with both similar and varied sizes offers greater potential for exploration and optimization within the HMoE model. More comprehensive and in-depth analyses are provided in the Appendix.

\subsection{In-depth Analyses on HMoE Experts}
\label{expert_analysis}

\begin{figure}[t]
    \centering
    \begin{subfigure}[b]{0.49\columnwidth} 
    \centering
    \includegraphics[width=\linewidth]{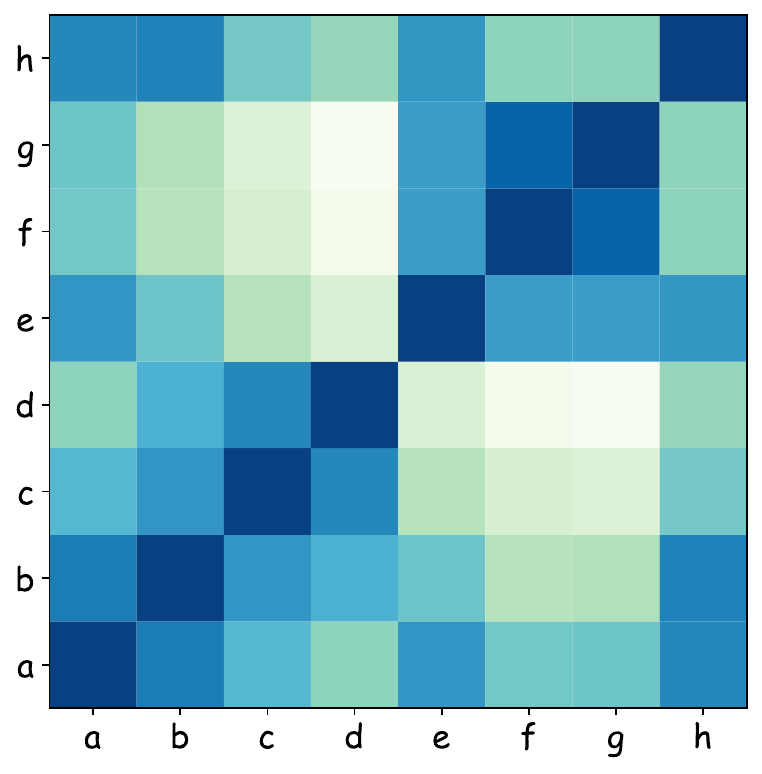} 
    \caption{Similarity Analysis}
\end{subfigure}
\hfill 
\begin{subfigure}[b]{0.50\columnwidth} 
    \centering
    \includegraphics[width=\linewidth]{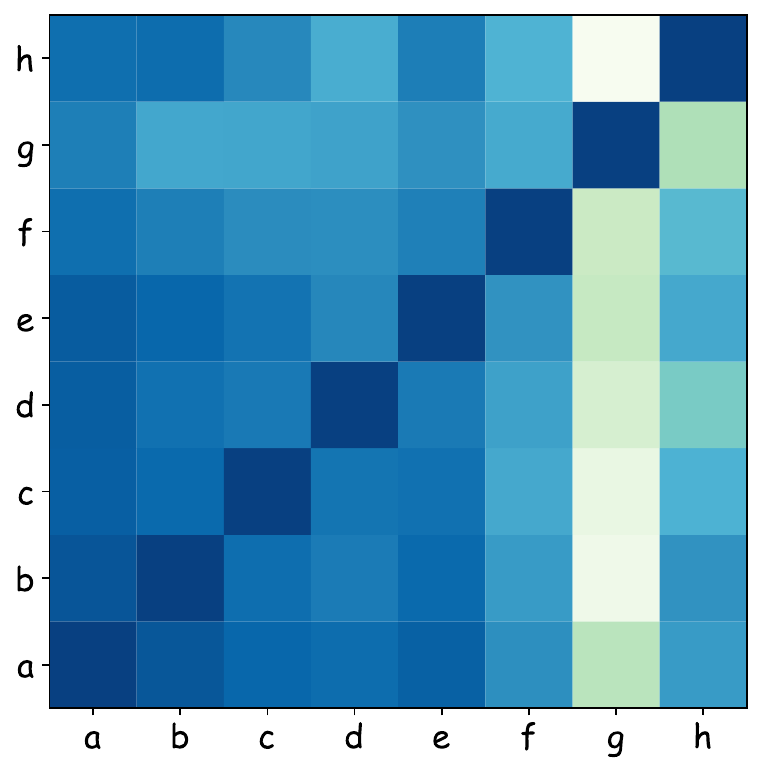}
    \caption{Synergy Analysis}
\end{subfigure} 
    \caption{Similarity and synergy analysis of HMoE's experts with the arithmetic strategy. The relative expert sizes are $\{9,11,13,15,17,19,21,23\}$ as experts from \textit{a} to \textit{h}.}
    \label{fig:expert_analysis}
\end{figure}

\begin{figure}[t]
    \centering
    \includegraphics[width=0.85\columnwidth]{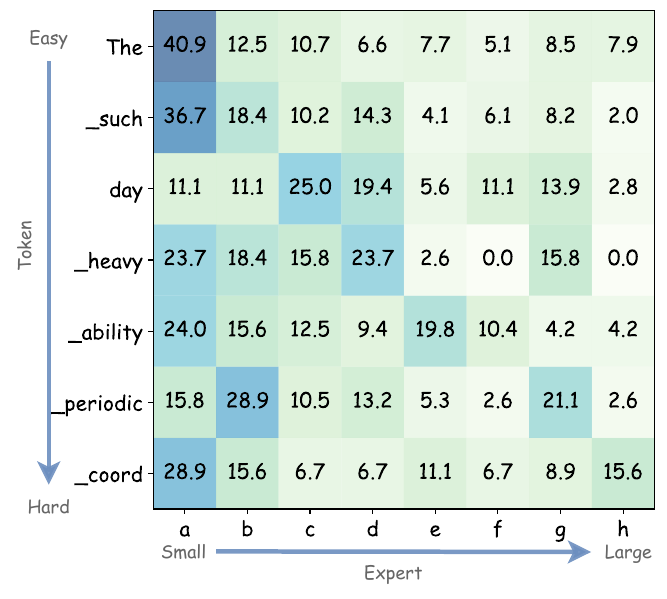}
    \caption{Visualization of activated experts ratio to tokens with different understanding difficulty. The expert size design is the same as Figure \ref{fig:expert_analysis}.}
    \label{fig:token_analysis}
\end{figure}

Figure \ref{fig:expert_analysis} (a) presents a similarity analysis of HMoE's experts of different sizes. Each heatmap cell represents the Wasserstein distance between token distributions of expert pairs on downstream tasks. We find that experts of similar sizes typically show greater similarity. Clustering is seen among experts with similar sizes (e.g., expert a/b, c/d, f/g). This indicates that experts with similar sizes tend to develop analogous capabilities, showing the significance of heterogeneity.

Figure \ref{fig:expert_analysis} (b) shows the synergy analysis among experts of different sizes. Each cell in the heatmap represents the KL divergence between token distributions of the x-axis and y-axis experts. Results indicate that smaller experts collaborate more than larger ones, while larger experts are more specialized. This suggests smaller experts in our HMoE have more generalized capabilities.

Figure \ref{fig:token_analysis} shows the activation ratios of experts for tokens with varying difficulty levels. The activation ratio is the frequency that a token activates each expert divided by the total activations. Complex tokens activate larger experts more often, while smaller experts are consistently activated due to their general capabilities. 

It is noteworthy that, although we present only a few examples, this phenomenon is universally observed. This suggests that our HMoE model effectively allocates tokens to appropriate experts.

\section{Related Work}

The Mixture of Experts (MoE) model was first proposed by \citet{adaptive_moe}, where each expert independently learns a subset of the complete dataset and is then integrated into a unified system. Building on this, \cite{smoe} introduced the Sparsely-Gated Mixture-of-Experts layer (SMoE), which employs a gating network for expert selection and proposes a top-K routing strategy, where a fixed number of experts are selected for each token. Further advancements were made by Gshard \cite{gshard} and SwitchTransformer \cite{switch_transformer}, which incorporated MoE into the Transformer architecture's Feed-Forward Network (FFN) layers, utilizing top-1 and top-2 routing strategies, respectively. Expert-choice MoE \cite{expert_choice_routing} introduced Expert Choice Routing, allowing each expert to independently select a certain number of tokens, thereby achieving perfect load balancing. AutoMoE \cite{automoe} establishes a search space tailored for small-scale heterogeneous MoE utilizing the top-1 routing strategy and employs Neural Architecture Search to derive a sub-network. Their experiments focus on machine translation tasks, and their approach is not suitable for pre-trained language models. \citet{not_all_experts_are_equal} illustrate that not all experts are equal in the MoE model. They discard less important experts and find the model that keeps the most performance. More recently, \cite{topp} introduced the top-P routing strategy, dynamically allocating the number of experts to each token. 
Our work is the first work exploring HMoE as a base language model based on top-K and top-P routing strategies. Diverse expert sizes in our HMoE inherently result in variances in expert proficiencies. Under the same average activation setting, our expert parameter allocation is more reasonable, ultimately achieving higher performance.

\section{Conclusion}

In this work, we propose a novel HMoE model, featuring experts of varying sizes to handle different token complexities. We enhance it by proposing a new training objective and exploring expert size distribution. Our experimental results show that HMoE improves both performance and computational efficiency. We believe that our work opens new avenues for the development of large language models. Future research could explore further optimization techniques and broader applications of heterogeneous expert architectures, potentially extending the benefits observed in this study to an even wider array of natural language processing tasks.

\bibliography{ref}

\appendix

\section{Limitation}

While our study highlights the substantial benefits of HMoE, several pathways for enhancement and exploration remain. Firstly, our initial experiments have yielded promising results, especially with increased training FLOPs. We anticipate even greater efficacy and scalability with larger datasets and models. Future work will focus on validating these effects on a larger scale and conducting more comprehensive analyses. Secondly, we have begun to explore the heterogeneity among experts. Although our current configurations have shown superior performance compared to traditional MoEs, we recognize the potential for discovering even more optimal setups. Future research will delve deeper into various configurations and routing strategies to identify the best solutions for diverse applications, thereby unlocking even greater performance and efficiency. Lastly, despite our optimized model and training processes achieving faster training speeds for HMoEs compared to traditional MoEs, there is still room for improvement, particularly in hardware adaptation. We believe that HMoE can achieve even faster training speeds with further optimization.

\section{Efficient Training of Heterogeneous MoE }

The efficient training of heterogeneous MoE models presents significant challenges to existing training approaches, necessitating innovative solutions to overcome these obstacles. 
One primary issue stems from the fact that experts do not have uniform shapes, which invalidates the traditional batched matrix multiplication method for expert computation. 
To address this challenge, Megablocks~\cite{gale2022megablocksefficientsparsetraining} implements efficient block sparse matrix multiplication kernels, which effectively handle the complexities introduced by variable-sized experts. 
Another concern is the problem of unbalanced computation and communication arising from the heterogeneous nature of experts, which can lead to inefficient resource utilization. 
To mitigate these issues, ES-MoE~\cite{es-moe} introduces expert-wise offoading and dynamic expert placement strategy.
This approach involves performing expert computation in a serialized manner. Expert parameters are offloaded to CPU memory and are fetched back to GPU memory as needed, based on the distribution of tokens.  
By doing so, ES-MoE not only reduces GPU memory overhead incurred by expert parameters but also alleviates the computation load imbalance issue, leading to better hardware resource utilization. 
Future research in the area may focus on developing more sophisticated load-balancing techniques and optimizing memory management strategies both for model states and activations.

\section{Detailed Model Setting}
\label{detailed_setteing}

All methods are based on the Transformer decoder-only architecture following LLaMa~\cite{llama1}. We employ the LLaMa2~\cite{llama2} tokenizer with a vocabulary size of 32,000.
We conducted a small-scale experimental exploration to determine the setting of model parameters.
For the Dense-0.4B model, we configure 12 Transformer Blocks, with the hidden dimensions of the FFN layers being 12,288. In the attention layer, we use 12 heads, each with a dimension of 64. For the Dense-1B model, we also configure 12 Transformer Blocks, but the hidden dimensions of the FFN layers are set to 32,768. In the attention layer, there are 16 heads, each maintaining a dimension of 64.

For both MoE (homogeneous MoE) and HMoE models, we utilize two different model sizes. Each layer in the MoE model contains 8 experts.
In the configuration with 0.4B total parameters, the total hidden dimension for all experts in each MoE layer sums to 12,288, and there are 12 Transformer Blocks. All other specifications align with Dense-0.4B settings.
In the configuration with 3B (2.55B) total parameters, the aggregate hidden dimension for all experts in each MoE layer is 32,768 and there are 12 Transformer Blocks. All other specifications match those of Dense-1B settings. For HMoE, the distribution of expert sizes follows an arithmetic progression.

For Homogeneous MoE, we set the load balancing loss coefficient to $1\times10^{-2}$, as implemented in \citet{topp}. For HMoE, we set the coefficient of parameter penalty loss as 0.1. For the Top-P routing strategy, we set the coefficient of router entropy loss as $3\times10^{-2}$.

\section{Detailed Training Setting}

Our models are trained utilizing NVIDIA A800 (80G memory) or H800 GPUs (80G memory). The AdamW optimizer is used, with a first-moment decay of $\beta_1=0.9$ and a second-moment decay of $\beta_2=0.999$. A weight decay of $1 \times 10^{-5}$ is applied. The learning rate is gradually increased from 0 to $1 \times 10^{-4}$ over the initial 1000 steps and is maintained thereafter. The context length is set to 4096, and the global accumulated batch size is 640. All experiments use a unified random seed value of 12345. We implemented the Zero2~\cite{deepspeed_zero} strategy to accelerate model training and gradient checkpointing to save GPU memory. All model and training code is developed with the torch \cite{pytorch} library.

\section{Detailed Introduction of MoE}

\subsubsection{Mixture of Experts}
Different from dense models, most MoE models replace the FFN layer of the transformer \cite{transformer} block with the MoE layer. The MoE layer consists of a router $g_i(\cdot)$ and multiple experts $\{e_1, e_2, ..., e_N\}$. The experts are composed of a set of independent Feed-Forward Network (FFN) layers. Experts are responsible for processing the input data according to their specialized knowledge. For each token, a subset of experts is activated to execute computations, and the router is responsible for generating a probability distribution. The probability of this distribution indicates the likelihood of assigning the token to each expert. We obtain the output of MoE layer based on following process:
\begin{equation}
\begin{split}
    \text{MoE}(\mathbf{x}) &= \sum_{i}^N g_i(\mathbf{x}) \cdot e_i(\mathbf{x}), \\
    e_i(\mathbf{x}) &= \text{FFN}_i(\mathbf{x}),
\end{split}
\end{equation}
where $\mathbf{x}$ is the input states of current layer.

\subsubsection{Routing Strategy}

The routing strategy is applied to select experts to be activated from $N$ experts. The \textbf{Top-K Routing}~\cite{topp} strategy is one of the most widely-used strategy, which always activates a fixed number of experts for each token. We first calculate the probability distribution $\mathbf{P}$ using a softmax function. $\mathbf{P}$ represents the initial score of selecting each expert. Then, we keep the highest $k$ scores and normalize them. The detailed computation is as:

\begin{equation}
    \mathbf{P} 
    = softmax(\mathbf{W_r} \cdot \mathbf{x})
    =\frac{\exp \left( \mathbf{W_r} \cdot \mathbf{x}  \right)}{\sum_{j=1}^N \exp \left (\mathbf{W_r} \cdot \mathbf{x} \right)},
\end{equation}
\begin{equation}
g_i(\mathbf{x})= \begin{cases}
\frac{P_i}{\sum_{j \in \operatorname{Top-K}(\mathbf{P})} P_j}, & i \in \operatorname{Top-K}(\mathbf{P}) \\
0, & i \notin \operatorname{Top-K}(\mathbf{P}),
\end{cases}
\end{equation}
where $\operatorname{Top-K}(\mathbf{P})$ returns the indices of the largest $k$ elements in $\mathbf{P}$, and $\mathbf{W_r}$ is a learnable router parameter.

Recently, \textbf{Top-P Routing}~\cite{topp} is proposed to dynamically activate different number of experts for each token. Specifically, we first obtain $\mathbf{\Tilde{P}}$ by sorting $\mathbf{P}$ from highest to lowest. Then given a fixed threshold $p$, which is a hyperparameter, if the highest probability is larger than threshold, we only use one expert. Otherwise, we progressively add additional experts until the cumulative probability exceeds the threshold $p$. The detailed computation is as:
\begin{equation}
t=\underset{k \in\{1 \ldots, N\}}{\operatorname{argmin}} \sum_{j<=k} \mathbf{\Tilde{P}}_{j} \geq p,
\end{equation}
\begin{equation}
\operatorname{Top-P}(\mathbf{P}) = \{\operatorname{Index}(1), ..., \operatorname{Index}(t)\},
\end{equation}
\begin{equation}
g_i(\mathbf{x})= \begin{cases}
\frac{P_i}{\sum_{j \in \operatorname{Top-P}(\mathbf{P})} P_j}, & i \in \operatorname{Top-P}(\mathbf{P}) \\
0, & i \notin \operatorname{Top-P}(\mathbf{P}),
\end{cases}
\end{equation}
where $t$ represents the minimum number of experts that need to be activated. $\operatorname{Index}(j)$ returns the indices of element $\mathbf{\Tilde{P}}_j$ in original distribution $\mathbf{P}$.

\section{Further Ablation on Expert Heterogeneity}

\begin{figure}[t]
    \centering
    \includegraphics[width=0.85\columnwidth]{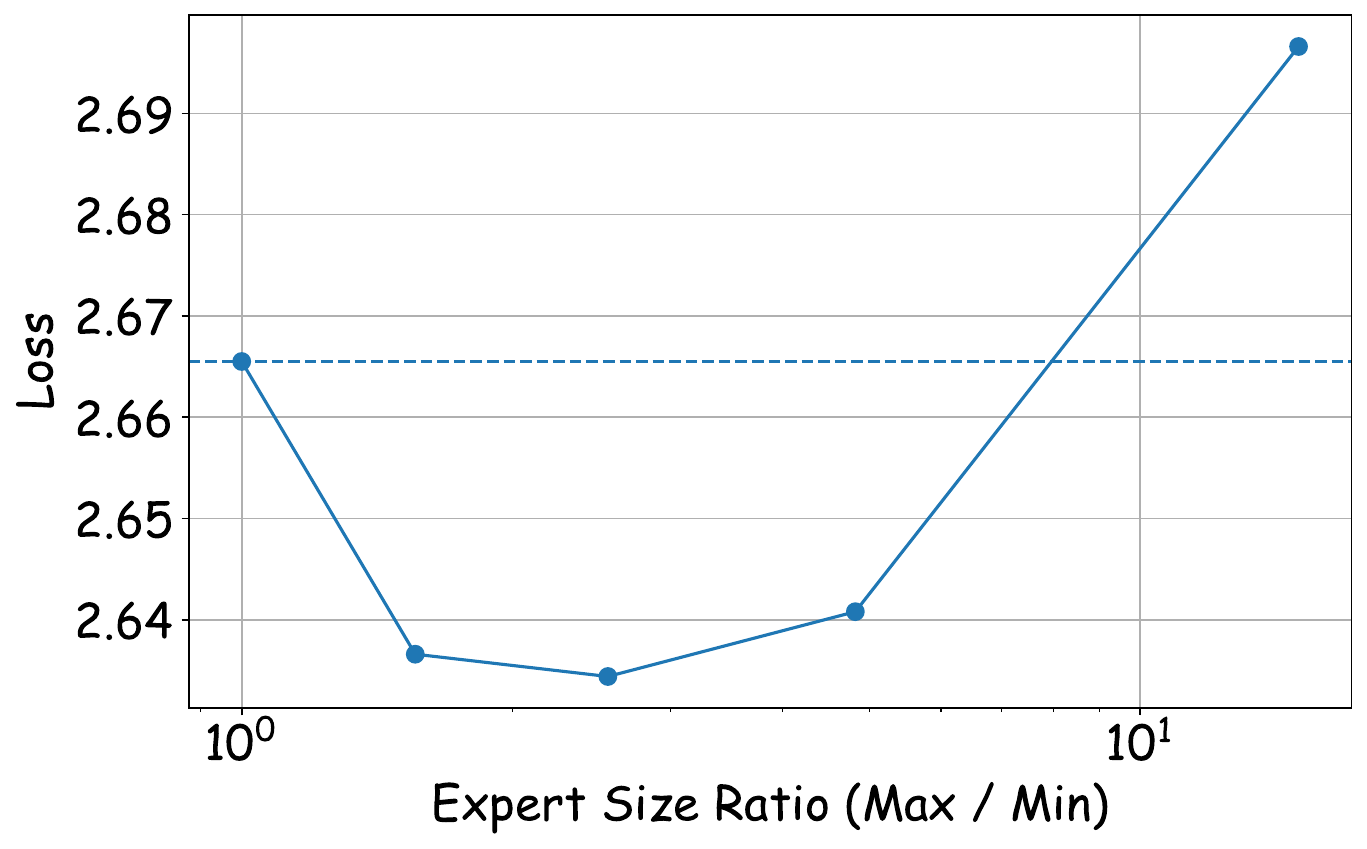}
    \caption{Various distributions of expert sizes in HMoE and their corresponding losses. All distributions follow arithmetic strategy. The x-axis represents the ratio of the size of the largest expert to the size of the smallest expert within the distribution.}
    \label{fig:heterogeneity}
\end{figure}

Our experiments reveal a strong correlation between loss and the performance of downstream tasks: lower loss generally leads to better performance. With this insight, we investigated how to determine Expert Heterogeneity. Figure \ref{fig:heterogeneity} illustrates the loss obtained by training HMoE using an arithmetic sequence strategy with varying levels of variance, all within the same computational budget. We observed that as the ratio between the largest and smallest experts increases (i.e., as the variance increases), the model's performance initially degrades but then improves. This suggests that in the heterogeneous design of HMoE, an optimal level of heterogeneity enhances performance compared to either excessive heterogeneity or complete homogeneity. This is consistent with the reason why the geometric distribution strategy has poor results. A large gap in expert ability is not conducive to model training and may lead to representation collapse. Based on these findings, we have adopted a relatively balanced heterogeneous distribution in our main experiment.

\section{Optimal Activated Model Parameters}

\begin{figure}[t]
    \centering
    \includegraphics[width=0.85\columnwidth]{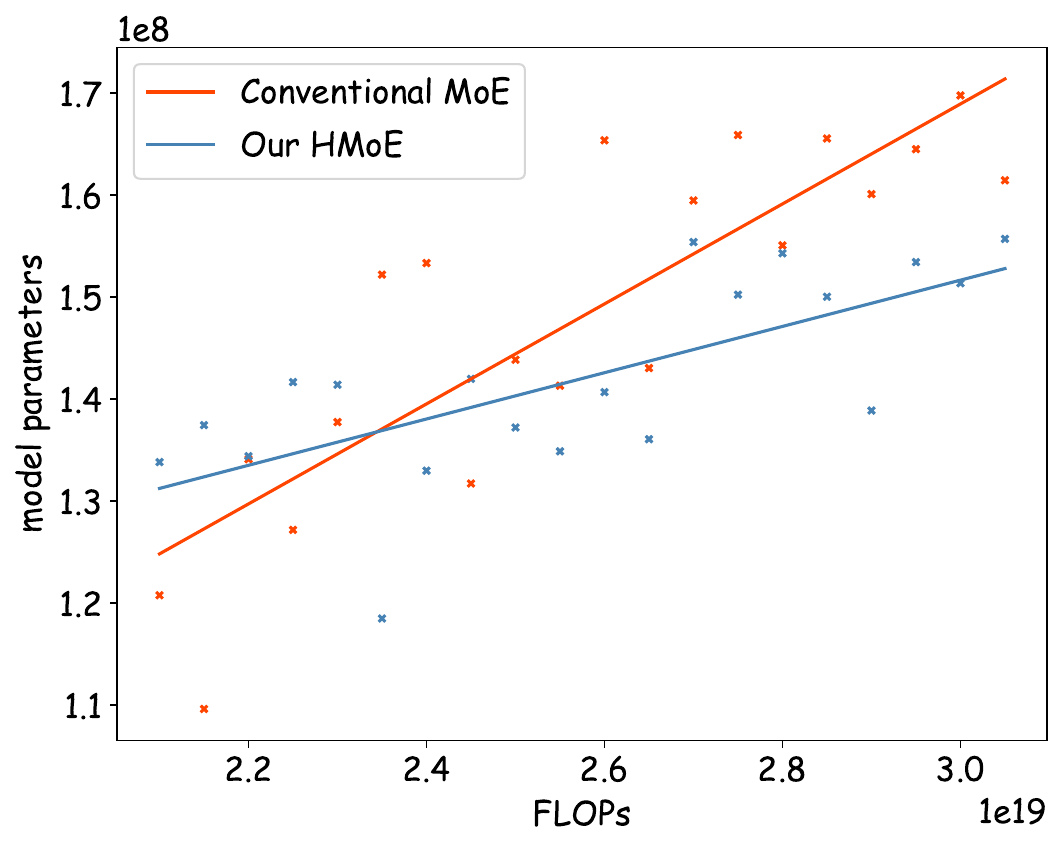}
    \caption{Optimal activated model parameters of our HMoE (Top-P) and conventional MoE (Top-P) under different training FLOPs.}
    \label{fig:isoflops_params}
\end{figure}

We recorded the activated parameters that yielded the lowest loss at different training costs. Figure \ref{fig:isoflops_params} illustrates that initially, the optimal number of activated parameters for the homogeneous MoE is lower than that for the HMoE. However, as the training FLOPs increase, the optimal number of activated parameters for the HMoE decreases. The crossover point occurs at approximately $2.4 \times 10^{19}$ FLOPs, which is relatively low for pre-training models. Considering the high computational costs associated with training modern large-scale models, this underscores the superior performance of HMoE as a base model for such training.

\section{Activated Parameter Ratio Analysis}

\begin{table}[t]
\centering
\begin{tabular}{l|c}
\toprule
\textbf{Task} & \textbf{Activated Parameter Ratio} \\
\midrule
ARC-Challenge   & 21.09 \\
ARC-Easy   & 20.23 \\
\bottomrule
\end{tabular}
\caption{Average Activated parameter ratios (\%) in HMoE layers for ARC~\cite{arc} tasks.}
\label{table:activated_ratios}
\end{table}

We present the activated parameter ratios of ARC tasks in HMoE layers in Table \ref{table:activated_ratios}. Specifically, we observe that ARC-Challenge activates more parameters compared to ARC-Easy. This implies that our model can dynamically activate parameters based on the difficulty of the task. This phenomenon is consistent with that in the MoE with Top-P routing strategy~\cite{topp}. By activating more parameters for more difficult tasks, the model achieves better performance, while for simpler tasks, it gains higher efficiency. This approach balances efficiency and performance. To be noted, the difference in activated ratios between difficult and simple tasks is not very large, ensuring stable computational costs.

\section{Expert Activation Patterns}

\begin{table}[t]
\centering
\begin{tabular}{c|p{5cm}}
\toprule
\textbf{Expert Dim} & \textbf{Top Tokens} \\
\midrule
2304 & the, such, your, these, most, you, both, no, they, each \\
3328 & tables, valley, sun, temper, places, day, war, water, through, clean \\
3840 & known, least, lowest, immediately, bare, heavy, known, higher, several, independent \\
5376 & \_ly, \_zen, \_icker, \_last, \_per, \_var, \_orous, \_next, \_end, \_flat \\
5888 & \_decom, \_iz, \_ro, \_inf, \_scra, \_coord, \_er, problem, \_och, \_foss\\
\bottomrule
\end{tabular}
\caption{Top activated tokens for each expert.}
\label{tab:expert_tokens}
\end{table}

We have recorded the tokens with the highest activation percentages for different sizes of experts in the ARC tasks. As shown in Table \ref{tab:expert_tokens}, smaller experts are most frequently activated by simpler words or words with less phonetic information. In contrast, larger experts are most frequently activated by suffix tokens. We believe that these suffix tokens are more ambiguous and thus more difficult to understand. Medium-sized experts, on the other hand, are more frequently engaged with tokens that have clearer semantics.

\section{Similarity Analysis}

\begin{figure}[t]
    \centering
    \begin{subfigure}[b]{0.49\columnwidth} 
    \centering
    \includegraphics[width=\linewidth]{figure/expert_analysis/similarity_analysis.pdf} 
    \caption{Heterogeneous Experts}
\end{subfigure}
\hfill
\begin{subfigure}[b]{0.50\columnwidth} 
    \centering
    \includegraphics[width=\linewidth]{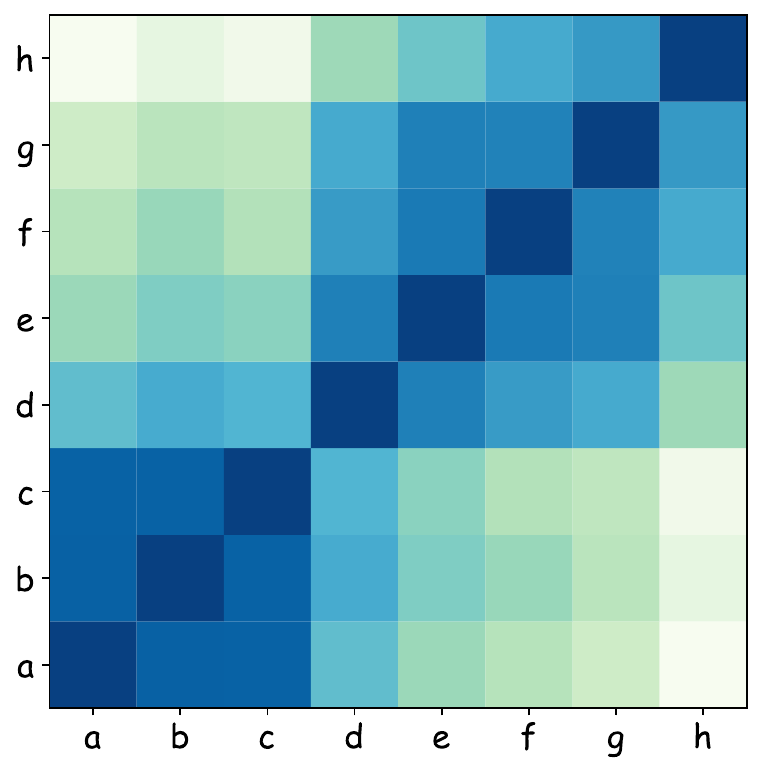}
    \caption{Homogeneous Experts}
\end{subfigure} 
    \caption{Similarity study of the heterogeneous and homogeneous experts. In the heterogeneous MoE, the relative expert sizes are $\{9,11,13,15,17,19,21,23\}$ as experts from \textit{a} to \textit{h}. In the homogeneous MoE, all experts have identical sizes.}
    \label{fig:expert_similarity_compare}
\end{figure}

We compared the behavior of experts in Heterogeneous MoE and Homogeneous MoE models. Figure \ref{fig:expert_similarity_compare} presents a similarity analysis of these experts, where each heatmap cell represents the Wasserstein distance between the token distributions of expert pairs on downstream tasks. In the Heterogeneous MoE setup, experts of similar sizes exhibit higher similarity. In contrast, in the Homogeneous MoE setup, where all experts are of equal size, we observed that experts tend to cluster into two groups. Specifically, experts a, b, and c display exceptionally high similarity. This comparison highlights the significant advantage of Heterogeneous MoE in facilitating expert differentiation compared to Homogeneous MoE.

\end{document}